\documentclass[runningheads]{llncs}
\usepackage{graphicx}
\usepackage{comment}
\usepackage{amsmath,amssymb} 
\usepackage{color}
\usepackage{hyperref}
\usepackage{svg}
\usepackage{subfig}

\begin{document}
\pagestyle{headings}
\mainmatter
\def\ECCVSubNumber{4567}  

\title{GAN-Based Facial Attractiveness Enhancement} 

\titlerunning{GAN-Based Facial Attractiveness Enhancement}
%
\author{Yuhongze Zhou
\and
Qinjie Xiao
}
%
\authorrunning{F. Author et al.}
%
\institute{Zhejiang University, Hangzhou, China} 
\maketitle

\begin{abstract}
We propose a generative framework based on generative adversarial network (GAN) to enhance facial attractiveness while preserving facial identity and high-fidelity. Given a portrait image as input, having applied gradient descent to recover a latent vector that this generative framework can use to synthesize an image resemble to the input image, beauty semantic editing manipulation on the corresponding recovered latent vector based on InterFaceGAN enables this framework to achieve facial image beautification. This paper compared our system with Beholder-GAN and our proposed result-enhanced version of Beholder-GAN. It turns out that our framework obtained state-of-art attractiveness enhancement results. The code is available at \url{https://github.com/zoezhou1999/BeautifyBasedOnGAN}.

\end{abstract}

\section{Introduction}
In the contemporary society, the pursuit for beauty has permeated people's daily life and a person with an attractive facial appearance can enjoy more compliments and opportunities compared to someone who is less facially attractive but other conditions are quite equal. Digital facial beautification has been highly demanded and brought out many applications, e.g. professional applications of retouching facial image for commercial use and social life need, supporting plastic surgery and orthodontics, selecting make-up and hair style, and beautifying applicants for certain screening situations like entertainment and modeling \cite{beautysurvey}.

Nowadays, beautifying facial images based on GAN is still an intriguing and relatively new topic that is worth delving into, because works in this aspect are relatively insufficient. Current introduced methods using GANs are \cite{diamant2019beholdergan,liu2019understanding}. 
Although these methods have opened a new field of vision in generating beautified images, they also raise challenging problems, such as image quality loss and keep-identity failure. CGANs dominated in these two methods which yields loss of synthesis resolution and quality. In spite of the fact that Beholder-GAN \cite{diamant2019beholdergan} sheds its light on the biological and sort of ingrained rationale of high beauty rating congruence over race, social class, gender and age \cite{beautysurvey}, through which, in results of Beholder-GAN, the higher a beauty score is, the higher possibility one person tends to feminize, rejuvenate and whiten, identity preserving in Beholder-GAN still cannot meet usual expectation about key identity attribute preservation, like sex. Therefore, we concluded that beautification of Beholder-GAN cannot control identity of generated images very well.

Major contributions in this paper include:
\begin{itemize}
\item We took advantage of high-resolution-and-quality image generation of StyleGAN \cite{karras2018stylebased} and semantic face editing finding by interpreting the latent space of GANs \cite{shen2019interpreting} to generate high-quality beautified images.
\item We managed to solve drawbacks in Beholder-GAN on the basis of it and compared our method with Beholder-GAN and modified Beholder-GAN in four respects, synthesized image realistic level, synthesized image quality, identity preserving level of generated images and the possibility of successful beautification.
\end{itemize}

Identity preservation here mainly means successful original image reconstruction and identity-preserving beautified image with quite high beauty level. Too much beautification usually pushes image away from original person identity, but beautified level can be controlled by parameters, which should not affect the evaluation of beautification. Our proposed approach achieves excellent performance in both high-quality and identity preserving.

\section{Overview}
The rest of the paper is organized as follows: Section 3 investigates previous works in facial beautification, facial attractiveness estimation and GANs providing foundation for beautification. Section 4 describes our approach. Evaluation experiments are detailed in Section 5. Section 6 presents further discussion and limitation. We concludes the paper in Section 7.

\section{Related Work}
\subsection{Popular Beautification Methods}
Traditional face beautification can be separated into 2D and 3D. Make-up and hair style transfer can also achieve excellent beautifying effect.
\subsubsection{2D Facial Image Beautification}
In facial image beautification, expect from some novel approaches, many achievements have been obtained in facial shape beautification and facial skin beautification \cite{Zhang2018}. In others words, image pixel color correction \cite{Zhang2018,2dbeauty1} and face geometric correction \cite{Li2015DeepFB,Data-drivenenhancementoffacialattractiveness,2dbeauty2} are prevailing and effective directions without affecting identity.

\cite{Zhang2018} improved multi-level median filtering method and performed filtering operation seven times on each facial image to remove facial blemishes after using the ASMs model to get facial landmarks. \cite{2dbeauty1} proposed a system that removes wrinkles and spots but preserves skin roughness using nonlinear filter.

\cite{Li2015DeepFB} trains the Deep Beauty Predictor (DBP) to capture potential relations between face shape (facial landmarks) and beauty score and then modifies facial landmarks of the original image to achieve beautification. In \cite{Data-drivenenhancementoffacialattractiveness}, it calculates distances among facial landmarks to form a ``face space'', searches a nearby point in this space to find point with higher beauty score and uses 2D warp to map input image to beautified output image.

Most methods mentioned above are on the basis of automatic beauty rating, considering the circumstance where such a subjective concept is beauty that data-driven methods for facial attractiveness estimation \cite{FacialAttractiveness:BeautyandtheMachine,beautyrater1,diamant2019beholdergan,beautyrater2} have aroused increasingly attention among communities of computer vision and machine learning. Technologies about automatic human-like beauty prediction can reduce investment of human power as well.
\subsubsection{3D Facial Beautification} 
3D facial beautification can be applied to more professional fields than 2D, like entertainment and medicine. \cite{3dbeauty1,3dbeauty2} both corrected facial asymmetry according to preference towards more symmetrical faces, but \cite{3dbeauty2} applied face proportion optimization by Neoclassical Canons and golden ratios as well. For medical beauty industry, there are also many developed systems \cite{medicalbeautytool} helping surgery planning and visualization effect.

\subsection{GANs Related to Beautification}

\subsubsection{Generative Adversarial Networks (GANs)} In recent years, Generative Adversarial Networks (GANs) \cite{goodfellow2014generative} have been explored extensively from lacking stability in the training process to capability of being more stable and generating any high-quality images, e.g. WGAN \cite{arjovsky2017wasserstein}, WGAN-GP \cite{gulrajani2017improved} and PGGAN \cite{karras2017progressive}, to more diversified and novel extended GANs, e.g. StyleGAN \cite{karras2018stylebased} that separates high-level facial attributes and stochastic variation automatically and unsupervisedly, and generates highly realistic images, a network mapping edge maps to colorized images \cite{isola2016imagetoimage}, and image-to-image translation \cite{choi2017stargan,zhu2017unpaired,liu2017unsupervised}.

\subsubsection{Conditional GANs (CGANs)}  CGANs have been used to generate images with certain feature or attribute, e.g. classified images \cite{grinblat2017classsplitting}, images with different age but the similar identity \cite{antipov2017face}, face pose synthesization \cite{FLAIRS1817643}, attributes editing operation \cite{perarnau2016invertible} and reconstructed animations based on facial expressions \cite{pumarola2018ganimation}. 

\subsubsection{Previous GAN-Based Facial Image Beautification}
\cite{diamant2019beholdergan} introduced Beholder-GAN that is based on previous works about Progressive Growing of GANs \cite{karras2017progressive} learning from low-resolution to high-resolution image and Conditional GANs (CGANs) \cite{mirza2014conditional} generating images conditioning on some attribute, e.g. class label, feature vector. Beholder-GAN uses a variant of PGGAN conditioned on a beauty score to generate realistic facial images. To be specific, it uses the SCUT-FBP5500 dataset \cite{liang2018scutfbp5500}, in which the 60-size beauty score distribution of each subject is reported, to train a beauty rater and uses this model to label the CelebAHQ dataset \cite{karras2017progressive} to enrich GAN training dataset. And recovery of latent vector and beauty score from input image and beautified image generation of trained Beholder-GAN are followed. \cite{liu2019understanding} quantifies correlations between beauty and facial attributes and extends StarGAN \cite{choi2017stargan} to transfer facial attributes to realize beautification.

\subsubsection{Image2LatentVector}
One critical part about identity preservation of GANs-generated image is to find one latent vector that is corresponding to a image similar to original one. In general, there are two approaches: 1) train an encoder that can map image to certain latent vector \cite{kingma2013autoencoding}; 2) initialize latent vector and optimize it using gradient descent \cite{lipton2017precise,abdal2019image2stylegan}. We follow the more popular and stable way, i.e. latent vector recovery.

\subsubsection{Image Attributes Editing}
One possible approach we can refer from previous works in beautification is to transform images between two domains, similar to CycleGAN \cite{zhu2017unpaired}, image-to-image translation \cite{liu2017unsupervised} and a extended multi-class image translation version, like StarGAN \cite{choi2017stargan}. Another is, unlike ways stated ahead that feed data containing information about certain classes/attributes into the network to learn, another viable method is to investigate into traditional GANs latent spaces and search some internal patterns between image semantic information and input latent vector \cite{shen2019interpreting}. 

\section{Approach}
From our experiments, in order to make a GAN generate images with variant beauty levels or one beautified image from an input original image, two main tasks are 1) recover/map a input original image to corresponding latent vector that can generate image resemble to the original image; 2) control of input of GAN related to beauty of generated image to realize beautification. The framework we describe here is StyleGAN and InterFaceGAN based beautification.
\subsection{InterFaceGAN}
A well-trained traditional GAN can be regarded as a deterministic function that can map a $z$, usually a d-dimensional latent vector in the Gaussian distribution and carrying semantic information, to a image, i.e. $I=G(z)$ and $s=f_s(I)$, where $I$ is generated image, $z\subseteq \mathbb{R}^d$, $z\sim \mathcal{N}(0,\mathcal{I}_d)$, $s\in S$, $S\subseteq \mathbb{R}^m$ is a semantic space with m attributes.

In \cite{karras2018stylebased}, while proposing the separability metric, it also mentioned an idea that it is possible to find direction vectors that correspond to individual factors of gradual change of image in a sufficiently disentangle latent space. There are some 2D linear latent code interpolation facial morphing works based on StyleGAN as well. However, \cite{shen2019interpreting} officially introduced an assumption that, for any binary attribute, there exists a hyperplane in the latent space considered as the separation boundary and semantics keep consistent while the latent code remains in one side of the hyperplane but turn to be the opposite in another side, and then empirically demonstrated and evaluated it. 

The ``distance'' defined in \cite{shen2019interpreting} is
$d(n,z)=n^{T}z$, $d\in[-\infty,\infty]$, where, considering a hyperplane with certain unit normal vector $n\in \mathbb{R}^d$, $z$ is sample latent code.

When a latent code ``vertically'' crosses from one side of a hyperplane corresponding to a certain binary attribute to the opposite, this attribute of image would vary accordingly from the negative to the positive with a high possibility and stability of other attributes preserving, i.e. $z_{edit}=z+\alpha n$, $f_s(G(z_{edit}))=f_s(G(z))+\lambda \alpha$. Besides, according to \emph{Property 2} mentioned by \cite{shen2019interpreting}, it is very likely that random latent codes from $\mathcal{N}(0,I_d)$ locate near enough to the given hyperplane. Therefore, images can be edited from original state to one containing certain positive attribute.

\subsection{StyleGAN and InterFaceGAN Based Beautification}
Given the state-of-the-art performance of StyleGAN, we decided to combine StyleGAN with InterFaceGAN. \cite{shen2019interpreting} successfully evaluated the correctness of the assumption of its framework that, for an arbitrary well-trained traditional GAN, there is a hyperplane in its latent space that can separate any binary attribute. We followed the novel single attribute manipulation in latent space \cite{shen2019interpreting} proposed and found that, in the latent space $\mathcal{W+}$ ($\mathcal{W}$ latent space with the truncation trick), beauty score, a continuous value, can be separated like a binary attribute.

\begin{figure}
\centering
\includegraphics{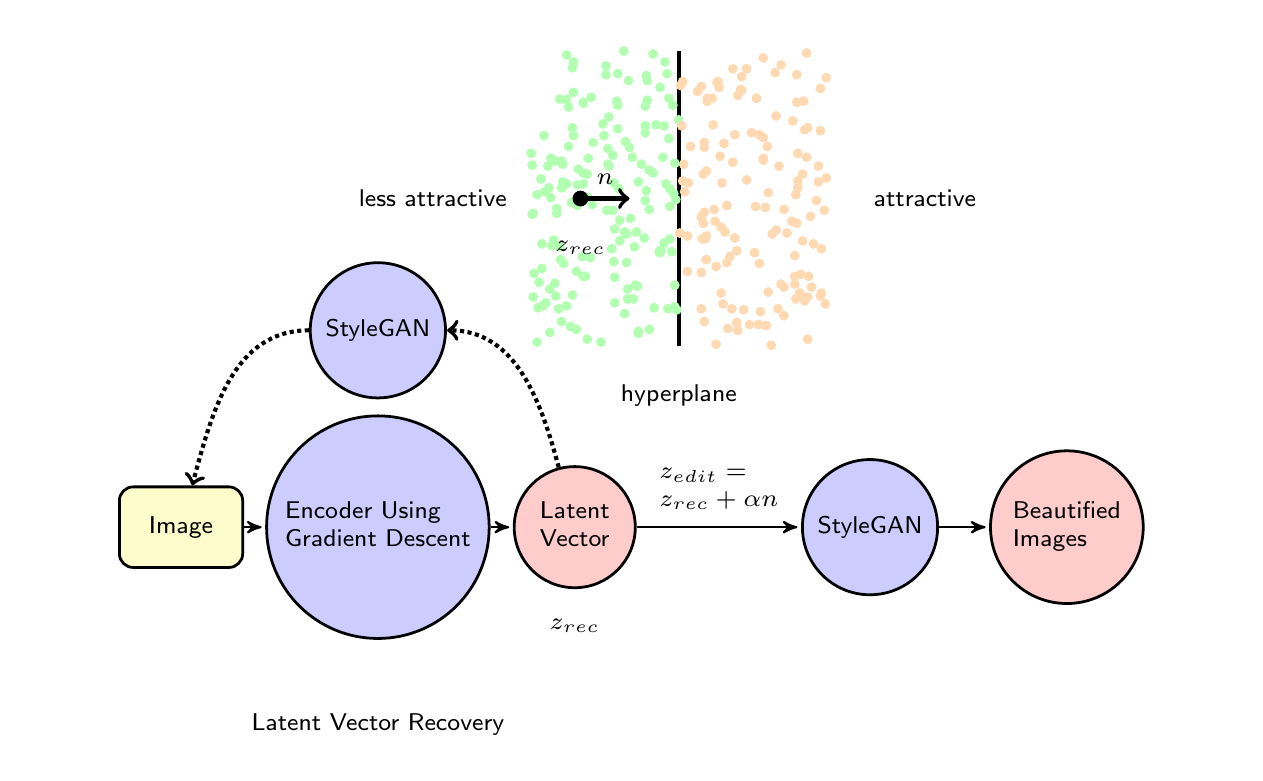}
\caption{Framework of StyleGAN and InterFaceGAN Based Beautification}
\label{fig:framework}
\end{figure}

\subsubsection{Generate hyperplane train dataset}
We randomly generated 40K sample images from StyleGAN as the dataset and stored their corresponding latent vectors and beauty scores, which are predicted by the same rating model in \cite{diamant2019beholdergan}.
\subsubsection{Train beauty hyperplane}
In the $\mathcal{W}$ latent space, we trained a linear SVM on beauty score with 5600 positive and negative samples separately and then evaluated them on the validation set (4800 samples) and the remaining set (The entire dataset consists of 40K random samples). Accuracy for validation set is $94.1250\%$ and for the remaining set is $68.7708\%$. The result showed that a beauty hyperplane exists in the latent space of StyleGAN and face editing \cite{shen2019interpreting} proposed is reasonable to be applied to beautification.
\subsubsection{Beautification}
We used the StyleGAN-Encoder \cite{styleganencoder} to get recovered latent vector from input image. The loss function used to optimize is as follows:
\begin{align}
L_{rec}=\lambda_{1}L_{pixel}+\lambda_{2}L_{vgg}+\lambda_{3}L_{ssim}+\lambda_{4}L_{percept}+\lambda_{5}L_{penalty}+\lambda_{6}L_{d}
\end{align}
\begin{align}
L_{pixel}=\frac{1}{W\times H}\sum_{w=1}^{W}\sum_{h=1}^{H}logcosh(x_{w,h}^{pred},x_{w,h}^{P})
\end{align}
\begin{align}
L_{vgg}=\frac{1}{N}\sum_{n=1}^{N}\left|F_{j}(x^P)-F_{j}(x^{pred})\right|
\end{align}
\begin{align}
L_{ssim}=\frac{1}{N}\sum_{n=1}^{N}(1-ssimMultiscale(x^{pred},x^{P}))
\end{align}
\begin{align}
L_{percept}=\frac{1}{N}\sum_{n=1}^{N}LPIPS(x^{pred},x^{P})
\end{align}
\begin{align}
L_{penalty}=\frac{1}{N}\sum_{n=1}^{N}\left|z^{pred}-z^{avg}\right|
\end{align}
\begin{align}
L_{d}=\frac{1}{N}\sum_{n=1}^{N}D(x^{pred})
\end{align}
\begin{align}
z_{i+1}=z_{i}-\eta\nabla_{z_{i}}L_{rec}
\end{align}
where $x^P, x^{pred}$ are the target and predicted images, $z^{pred}, z^{avg}$ are the estimated and average latent vectors in one batch, respectively, $F_j$ is the feature output of the jth VGG-16 layer, $ssimMultiscale$ is Multi-Scale Structural Similarity (MS-SSIM), $LPIPS$ is the metric proposed by \cite{zhang2018unreasonable} and the discriminator output for the predicted image is regarded as $L_d$. And then face editing was applied with a start ``distance'' of 0.0 and an end ``distance'' of 3.0.

\section{Experiments}
In this section, we compared our framework with current method, Beholder-GAN, and enhanced version of it, which, to some extent, solved identity-preserving weakness of Beholder-GAN.
\subsection{Enhanced Version of Beholder-GAN}
\subsubsection{A GAN conditioned on a beauty score and identity feature}
This version of Beholder-GAN is inspired by identity-preserving GANs, using facial feature as the input of the generator and proposing their own identity preserving loss functions \cite{facidgan,facerotation}. Based on Beholder-GAN, which constructed a generator conditioned on a beauty score, we added identity feature as another condition and reconstructed the loss function of the generator and output of the discriminator. For enriching training dataset, following \cite{diamant2019beholdergan}, we used its beauty rating model and FaceNet \cite{Schroff_2015} to label the FFHQ dataset with beauty scores and identity features.
\begin{align}
x =G(z\mid \alpha, \beta)
\end{align}
where $\alpha \in [0, 1]$ denotes the beauty score rating vector, $\beta$ is a normalized 512D feature embedding extracted from FaceNet, $z\sim \mathcal{N}(0,\mathcal{I})$ is a latent vector in random Gaussian distribution and $x$ is the generated image. The generator loss function is
\begin{align}
L_g=L_{adv}+\lambda L_{ip}
\end{align}
\begin{align}
L_{adv}=\frac{1}{N}\sum_{n=1}^{N}-D(G(z\mid \alpha, \beta))
\end{align}
\begin{align}
L_{ip}=\frac{1}{N}\sum_{n=1}^{N}\sum_{y}F(x^P)\log_2 F(x^{pred})
\end{align}

where $L_{adv}$ means an adversarial loss to distinguish real images from synthesized fake images \cite{wangp}, $L_{ip}$ denotes an identity preserving loss and its weight, i.e. $\lambda$, is 0.5 in our network. Preserving the identity in the training process is one key part in our method except following part of identity-preserving image reconstruction from input image. In this work, we referred from the identity preserving loss that is original exploited by \cite{facerotation} and made it calculate cross entropy using prediction from FaceNet fed with real and fake images respectively.

Aside from original version of Beholder-GAN making the discriminator predict beauty score as well as the real vs. fake probability, we reconstructed the PGGAN structure of the discriminator to fit additional feature condition and also applied the $l_2$ loss $(\gamma - \hat{\gamma})^2$ as the score estimated controller on the discriminator loss, where $\gamma$ is the concatenation of beauty scores and identity feature.
\begin{align}
    (\hat{P}(real),\hat{\alpha},\hat{\beta})=D(x)
\end{align}

\subsubsection{Beautification}
On the foundation of enhanced Beholder-GAN, which can learn beauty through input mixed label, another modification is more stable and correct face reconstruction from input image in spite of larger dimensional label input and more parameters.

We initialized input of gradient-descent latent vector recovery with ResNet-estimated $z_0$ and random $\alpha_0, \beta_0$, using an aggregate of the below separate loss functions as the final objective function to optimize for image-corresponding latent vector recovery with constant learning rate.
\begin{align}
L_{rec}=\lambda_{1}L_{pixel}+\lambda_{2}L_{vgg}+\lambda_{3}L_{ssim}+\lambda_{4}L_{percept}+\lambda_{5}L_{label}
\end{align}
\begin{align}
L_{label}=\frac{1}{N}\sum_{n=1}^{N}\left|l^{pred}-l^{P}\right|
\end{align}

where $x^P, x^{pred}$ are the target and predicted images with $l^P, l^{pred}$ as their corresponding labels. The target constant label is estimated by the FaceNet and the beauty rater we trained. We used the method similar to the stochastic clipping proposed by \cite{lipton2017precise} to clip labels to certain range every descent.

After recovering the latent vector $\hat{z}$, beauty score $\hat{\alpha}$ and facial feature $\hat{\beta}$, we fed recovered fixed $\hat{z},\hat{\beta}$ and continuously increased $\alpha_+$, using the previous recovered $\hat{\alpha}$ as its baseline, into the feed forward model $G(\hat{z}\mid \alpha_+, \hat{\beta})$, where $\alpha_+$ is an higher beauty level $(\alpha_+ \geq \hat{\alpha})$, to get beautified facial images.

The combination of these modifications endow the GAN with the ability to keep basic personal attributes, especially during face reconstruction and while a quite high beauty score is pushing the generated image away from person's identity.

\subsection{Evaluation}
\begin{figure}[ht]
  \subfloat[Method \raisebox{.5pt}{\textcircled{\raisebox{-.9pt} {1}}}]{
	\begin{minipage}[t][2.05cm][t]{
	   0.3\textwidth}
	   \centering
      \includegraphics[height=2cm]{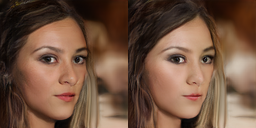}
	\end{minipage}}
 \hfill 	
  \subfloat[Method \raisebox{.5pt}{\textcircled{\raisebox{-.9pt} {5}}}]{
	\begin{minipage}[t][2.05cm][t]{
	   0.3\textwidth}
	   \centering

	  \includegraphics[height=2cm]{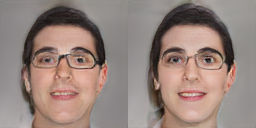}

	\end{minipage}}
 \hfill	
  \subfloat[Method \raisebox{.5pt}{\textcircled{\raisebox{-.9pt} {6}}}]{
	\begin{minipage}[t][2.05cm][t]{
	   0.3\textwidth}
	   \centering

      \includegraphics[height=2cm]{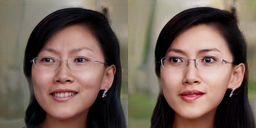}

	\end{minipage}}
\caption{Beautification-Disagreed Examples of Three Methods from Online Survey (The Right is the Beautified One)}
\label{fig:onlinesurvey}
\end{figure}

\begin{figure}[!ht]
  \subfloat[Method \raisebox{.5pt}{\textcircled{\raisebox{-.9pt} {8}}}]{
	\begin{minipage}[t][4.025cm][t]{
	   1\textwidth}
	   \centering

      \includegraphics[height=2cm]{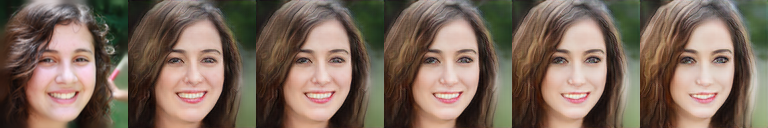}
       \includegraphics[height=2cm]{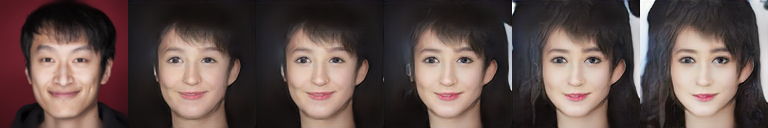}

	\end{minipage}}
	 \hfill	
  \subfloat[Our Approach(Method \raisebox{.5pt}{\textcircled{\raisebox{-.9pt} {9}}})]{
	\begin{minipage}[t][4.025cm][t]{
	   1\textwidth}
	   \centering

       \includegraphics[height=2cm]{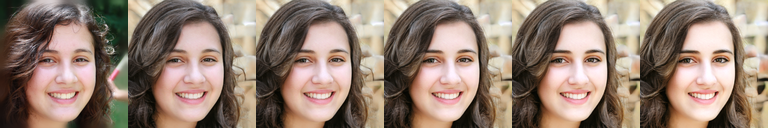}
    \includegraphics[height=2cm]{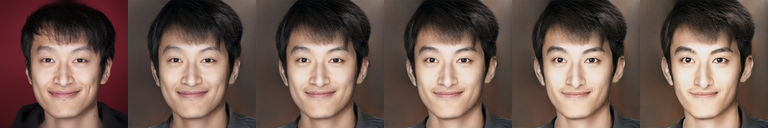}
	\end{minipage}}
	
\caption{Other Selected Six-Method Results}
\label{fig:results2}
\end{figure}

\begin{table}
\begin{center}
\caption{Comparison of Fr\'echet Inception Distance (FID), Blind/Referenceless Image Spatial Quality Evaluator (BRISQUE) of generated images, OpenFace Feature $L_2$ Distance (OFD) between the input and recovery images and Rating Agreement (RA) of our online survey among Beholder-GAN (BGAN), Modified Beholder-GAN (MBGAN) and StyleGAN \& InterFaceGAN Based Beautification (Our Approach) (OB: Original Input Image BRISQUE; RB: Result Image BRISQUE; B-IP: BRISQUE Incremental Percent)}
\label{table:table1}
\begin{tabular}{lllllll}
\hline\noalign{\smallskip}

Method & FID &\multicolumn{3}{c}{BRISQUE} & OFD& RA\\
\cline{3-5}
    & &OB & RB & B-IP & \\
\noalign{\smallskip}
\hline
\noalign{\smallskip}
\raisebox{.5pt}{\textcircled{\raisebox{-.9pt} {1}}}BGAN$^{\dagger\bigstar\bigstar}$ & 371.8965 & 7.0443 & 4.9705 & 29.4391\% & \textbf{1.3456} &N/A\\

\raisebox{.5pt}{\textcircled{\raisebox{-.9pt} {2}}}BGAN+$L_{rec}^{\dagger\bigstar\bigstar}$ & 371.8965 & 7.0443& 12.0087 & -70.4754\%& 1.0958& N/A\\

\raisebox{.5pt}{\textcircled{\raisebox{-.9pt} {3}}}BGAN+$L_{rec}$,w/o $L_{label}^{\dagger\bigstar\bigstar}$ & 371.8965 & 7.0443&11.4624 & -62.7197\%& \textbf{0.9900} & N/A\\

\raisebox{.5pt}{\textcircled{\raisebox{-.9pt} {4}}}BGAN$^{\dagger\star}$ & 86.0071 & 17.2466 & 6.9791 & 59.5334\% & 1.3392&90.1944\%\\

\raisebox{.5pt}{\textcircled{\raisebox{-.9pt} {5}}}BGAN+$L_{rec}^{\dagger\star}$ & 86.0071 & 17.2466& 15.1309 & 12.2675\%& 1.2118& N/A\\

\raisebox{.5pt}{\textcircled{\raisebox{-.9pt} {6}}}BGAN+$L_{rec}$,w/o $L_{label}^{\dagger\star}$ & 86.0071 & 17.2211&14.1685 & 17.7259\%&1.0059 & N/A\\

\raisebox{.5pt}{\textcircled{\raisebox{-.9pt} {7}}}MBGAN$^{\dagger\star}$ & 123.9326 & 17.2211& 22.5412 & -30.8931\%& 0.9342 & N/A\\
\raisebox{.5pt}{\textcircled{\raisebox{-.9pt} {8}}}MBGAN$^{\dagger\star\star}$& 172.2134
& 17.2211& 12.8857 & 25.1745\%&\textbf{0.8912} & \textbf{94.7667\%}\\

\raisebox{.5pt}{\textcircled{\raisebox{-.9pt} {9}}}Our Approach$^\ddagger$& \textbf{4.4159}\cite{karras2018stylebased} & 9.2128& 8.9666 & 2.6717\% & \textbf{0.1299} & 90.9583\% \\
\raisebox{.5pt}{\textcircled{\raisebox{-.9pt} {10}}}Our Approach$^{\dagger\bigstar}$& 9.1837  & 17.2211& 19.3427 & -12.3202\% & 0.1571 & N/A\\
\hline
\end{tabular}
\\
\footnotesize{$^\dagger$ 128$\times$128 resolution, $^\ddagger$ 1024$\times$1024 resolution, $^{\star}$ 12000K, $^{\star\star}$ 6011K, $^{\bigstar}$ 8908K, $^{\bigstar\bigstar}$ 6001K (the number of images used for training)}\\
\end{center}
\end{table}
For evaluation, in this paper, we calculated the FIDs using the same method in \cite{karras2018stylebased} (random selection of 50,000 images from the training set and the lowest distance encountered during training reported). We randomly chose 400 images as test input images to calculate BRISQUE using the ImageMetrics part of \cite{spmallick_2019} and OpenFace FD using \cite{amos2016openface}. All quantitative evaluations are represented in Table~\ref{table:table1}. To verify the correctness of relationship between beauty score/level and output beautified effect, we made use of the results from these 400 images to conduct an online user survey with 400 paired two-image contents with the distance of 0.1 in their beauty score $\alpha$ in both our reproduced Beholder-GAN and modified version, and the ``distance'' of 1.2 from the trained beauty hyperplane in our proposed approach, respectively. The raters were asked to point out samples showing failures of our methods. Fig.~\ref{fig:onlinesurvey} shows a few examples where the raters disagreed with the results of these beautification models. In Fig.~\ref{fig:results} and \ref{fig:results2}, the first image in the same row is the input image, and following are images with 0.05 beauty score and 0.3 ``distance'' increment from the recovered beauty score and the beauty hyperplane, from left to right, respectively for Beholder-GAN based and our method.

Images used by Method \raisebox{.5pt}{\textcircled{\raisebox{-.9pt} {1}}}-\raisebox{.5pt}{\textcircled{\raisebox{-.9pt} {3}}} are not aligned twice, while Method \raisebox{.5pt}{\textcircled{\raisebox{-.9pt} {4}}}-\raisebox{.5pt}{\textcircled{\raisebox{-.9pt} {10}}} uses image alignment. Image Alignment would make image quality lower a bit.
\section{Discussion and Limitation}
Although all issues stated below can be controlled and avoided, to some extent, by parameters, improvements in those aspects can make beautification algorithm robuster and more practical.
\subsection{Discussion}
In general, there have two ways to beautification based on GAN. One is CGAN-based approach and another is to use traditional GAN, which are rooted at different mechanisms.

In the experiments, when variants of PGGAN reach its maximum resolution, the longer it is trained, the better the FID scores become, but it is more likely for overfitting of GANs to happen. Despite FID score fluctuations, the relative relationship of FID scores among these models should be reflected through these evaluation experiments. The larger the dimension of label (not one-hot coding variables) added to CGAN is, the worse the realisticness and quality of generated images are, which can be verified by our evaluation and results, and can also be considered as the weakness of CGANs, while traditional GANs can synthesize more realistic images. We found that the combination of $L_{ip}$ and $L_{rec}$ based on Beholder-GAN, i.e. MBGAN, can achieve better results compared to only usage of $L_{rec}$ on Beholder-GAN which can improve the face reconstruction process.

During modification of Beholder-GAN, one issue we encountered is that, due to the usage of PGGAN, different from \cite{facerotation}, the initial generated images are blurry which somehow weakens the functionality of our generator loss function in MBGAN and also scales down the number of face recognition networks we can choose from, which requires the less sensitiveness of recognition networks towards pose and quality of facial images. Replacing network structure with another one that can work around these issues is a way for improvement as well as finding more informative identity features as the input of the generator.

Besides, one common pattern in these beautifications is that beautification in skin and makeup aspects comes first and then combination of skin and makeup beautification and facial geometric modification like face shape and the five sense organs is following with incremental beauty score and positive distance from the beauty hyperplane. The empirical thresholds for this in our experiment are 0.1 and 1.2 for Beholder-GAN based and our method, respectively. And within these thresholds, beautified images contain more realistic and practical value.

\subsection{Limitation}
For the image recovery results, it is hard for networks to recover detailed accessories in facial images (Fig.~\ref{fig:results}). In order to improve this, more diverse and comprehensive accessories-contained high quality face databases are required.

\begin{figure}
\centering
\includegraphics[height=5cm]{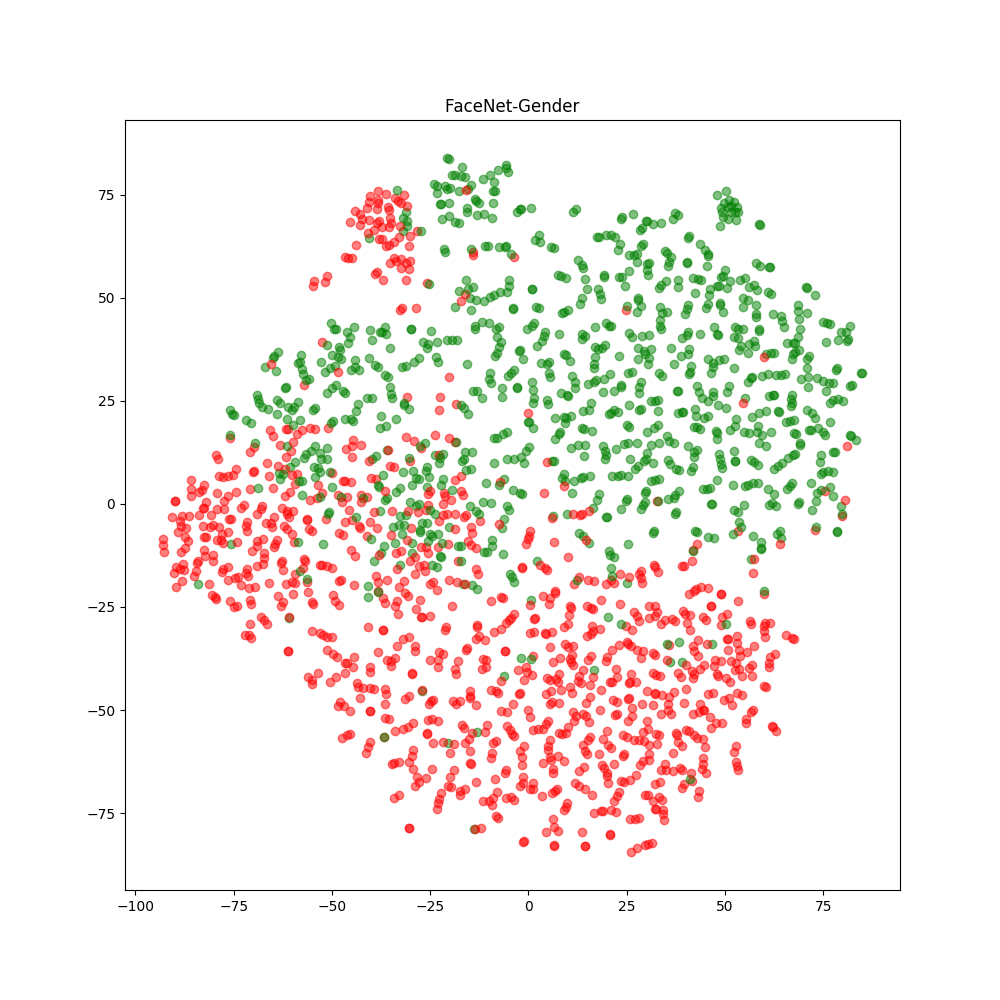}
\includegraphics[height=5cm]{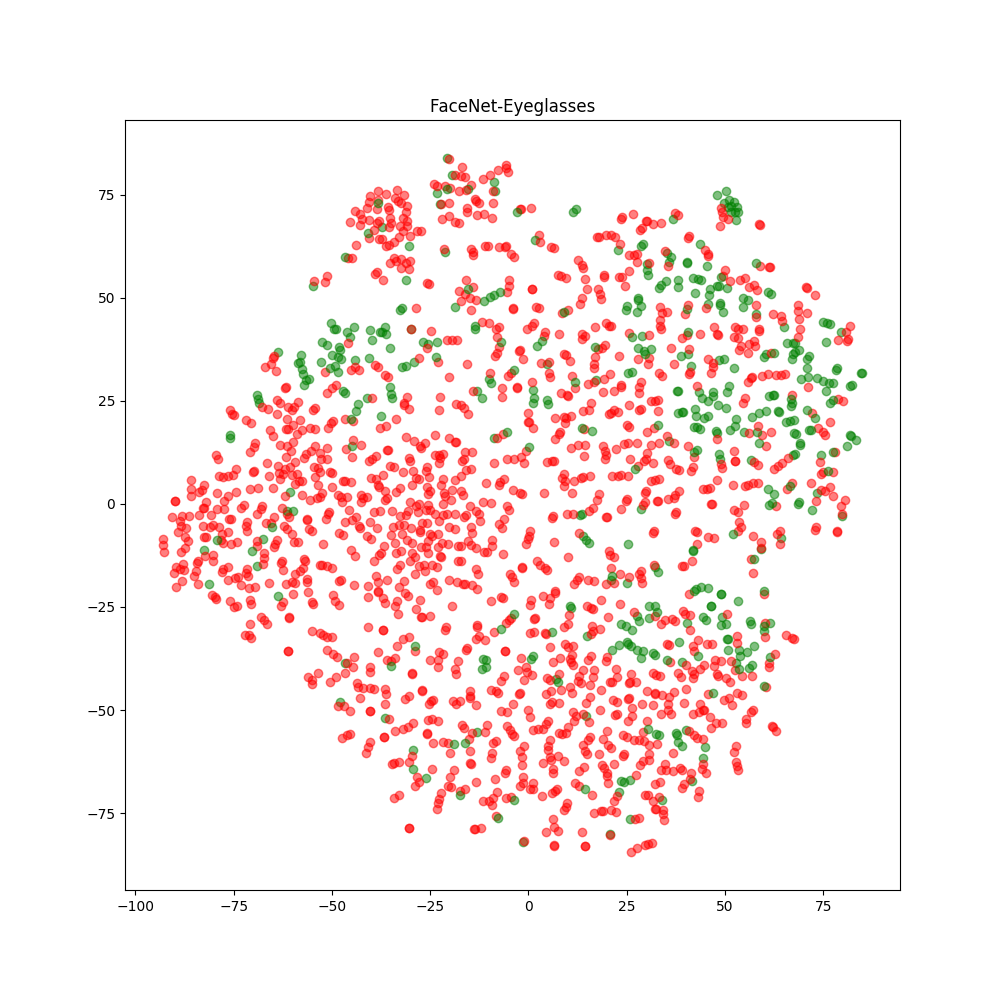}
\caption{FaceNet feature space of 2000 images from the FFHQ datasets. Each color represents a different attribute a person has in one plot. The attribute for the left plot is gender; for the right plot is wearing eyeglasses.}
\label{fig:featurevisualization}
\end{figure}

And unlike the advantage the latent space of StyleGAN possesses, how to find a way to truly make condition variables control information on generated images of CGANs is still a valuable problem that can be digged into. From our experiment results and related results from other papers, we found $\mathcal{W+}$ latent space of StyleGAN has excellent performance on binary attribute distinguishment, while Beholder-GAN and modified version of it still lose some control in various important facial features of original images during beautification process, like gender and eyeglasses. For example, encountering facial images with eyeglasses, in spite of successful recovery of eyeglasses, modified Beholder-GAN fails to maintain eyeglasses attribute with increasing beauty score compared to our approach (Fig.~\ref{fig:results}), although these aspects of modified Beholder-GAN might outperform Beholder-GAN a little bit. We assume this disadvantage results from the incapability of input identity features to separate obvious facial attributes from each other significantly. In order to discover the relationship between the input identity features and binary attributes, we visualize some features using t-SNE \cite{Maaten2008VisualizingDU} (Fig.~\ref{fig:featurevisualization}). We can see FaceNet cannot separate those features very well, which might give rise to attribute changing with increased beauty score. Current well-performed face recognition networks are driven by mass data, therefore, informative and binary-attribute-separated features that can represent any face properly, an analogy of traditional GAN's generating any face and one hyperplane likely found in traditional GAN's latent space separating one binary attribute, demands very comprehensive and challenging face datasets. The one-hot coding of attributes as conditional variables for well-labeled face datasets might be a circumvention that is worth trying.

Another thing we discovered is that high resolution of StyleGAN can produce recovered images of higher sharpness and quality than low resolution of it, in spite of the fact that both can achieve excellent identity preservation.

\begin{figure}[ht]
\subfloat[Beholder-GAN(Method \raisebox{.5pt}{\textcircled{\raisebox{-.9pt} {1}}})]{
	\begin{minipage}[t][8.1cm][t]{
	   1\textwidth}
	   \centering
	    \includegraphics[height=2cm]{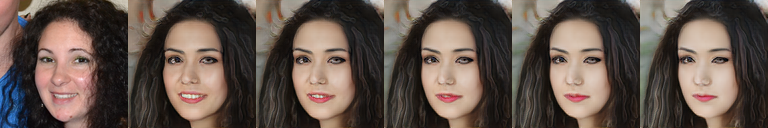}
	     \includegraphics[height=2cm]{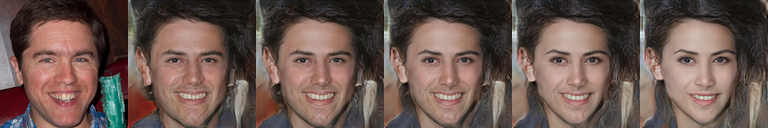}
      \includegraphics[height=2cm]{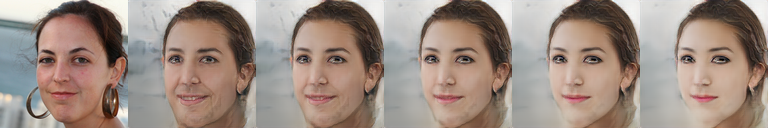}
      \includegraphics[height=2cm]{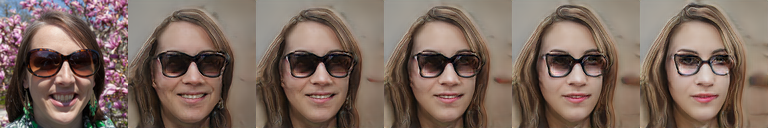}
	\end{minipage}}
 \hfill 	
   \subfloat[Method \raisebox{.5pt}{\textcircled{\raisebox{-.9pt} {3}}}]{
	\begin{minipage}[t][8.1cm][t]{
	   1\textwidth}
	   \centering
      \includegraphics[height=2cm]{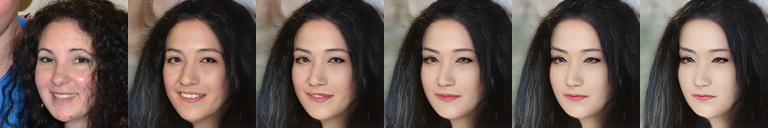}
      \includegraphics[height=2cm]{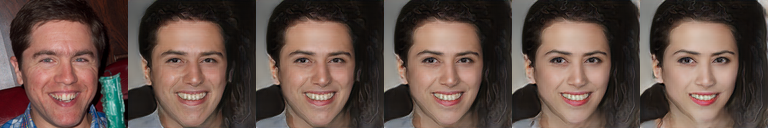}
      \includegraphics[height=2cm]{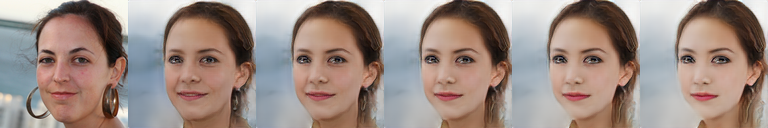}
      \includegraphics[height=2cm]{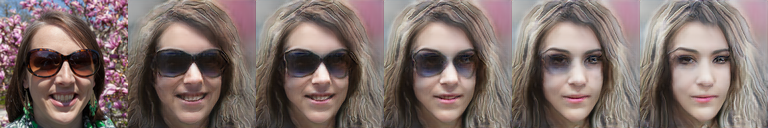}

	\end{minipage}}
	\end{figure}
	
	\newpage
	\begin{figure}[!ht]\ContinuedFloat
\subfloat[Beholder-GAN(Method \raisebox{.5pt}{\textcircled{\raisebox{-.9pt} {4}}})]{
	\begin{minipage}[t][8.1cm][t]{
	   1\textwidth}
	   \centering
	    \includegraphics[height=2cm]{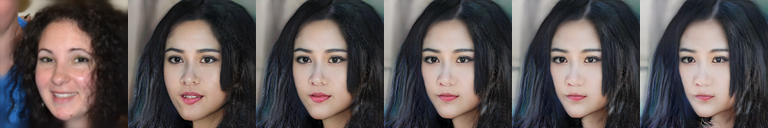}
	     \includegraphics[height=2cm]{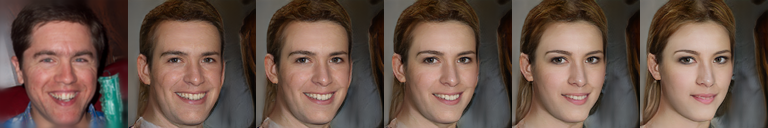}
      \includegraphics[height=2cm]{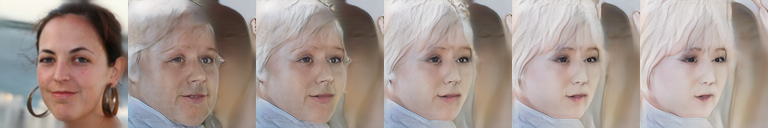}
      \includegraphics[height=2cm]{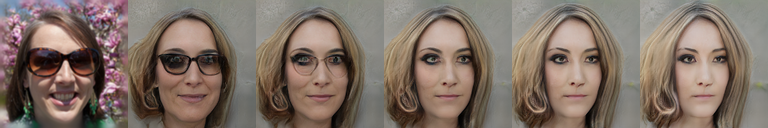}
	\end{minipage}}
 \hfill 	
   \subfloat[Method \raisebox{.5pt}{\textcircled{\raisebox{-.9pt} {6}}}]{
	\begin{minipage}[t][8.1cm][t]{
	   1\textwidth}
	   \centering
      \includegraphics[height=2cm]{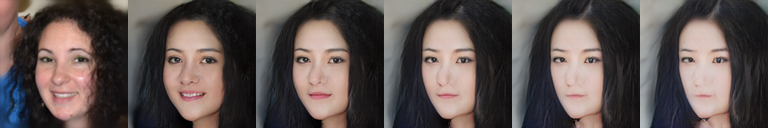}
      \includegraphics[height=2cm]{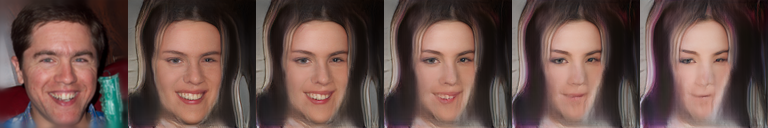}
      \includegraphics[height=2cm]{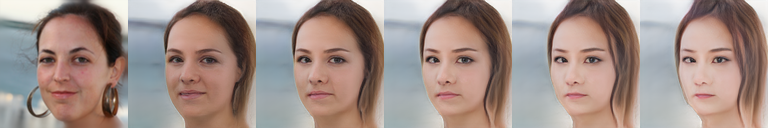}
      \includegraphics[height=2cm]{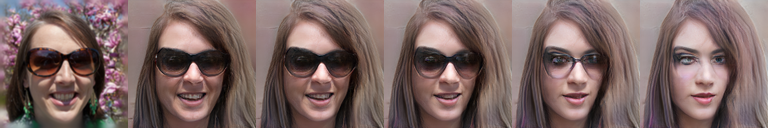}

	\end{minipage}}
	\end{figure}
	
	\newpage
	\begin{figure}[!ht]\ContinuedFloat
  \subfloat[Method \raisebox{.5pt}{\textcircled{\raisebox{-.9pt} {8}}}]{
	\begin{minipage}[t][8.1cm][t]{
	   1\textwidth}
	   \centering

      \includegraphics[height=2cm]{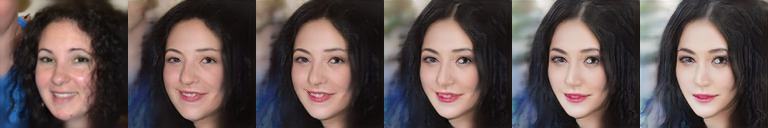}
       \includegraphics[height=2cm]{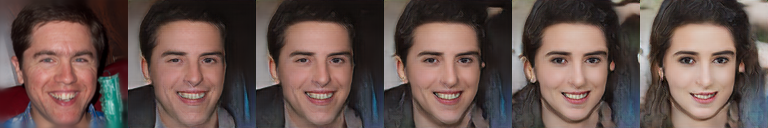}
      \includegraphics[height=2cm]{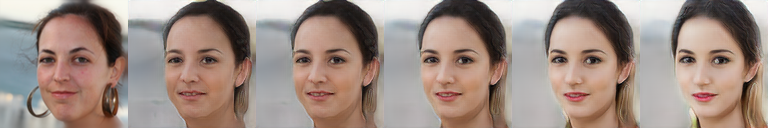}
      \includegraphics[height=2cm]{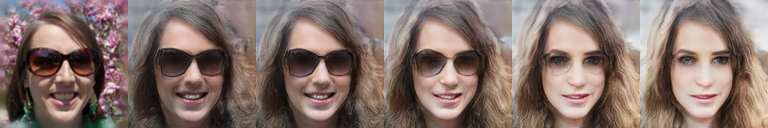}

	\end{minipage}}
	 \hfill	
  \subfloat[Our Approach(Method \raisebox{.5pt}{\textcircled{\raisebox{-.9pt} {9}}})]{
	\begin{minipage}[t][8.1cm][t]{
	   1\textwidth}
	   \centering

       \includegraphics[height=2cm]{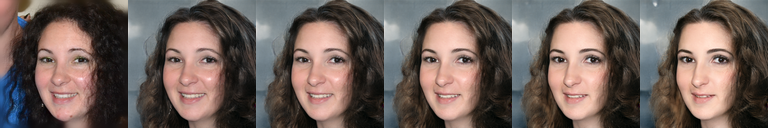}
    \includegraphics[height=2cm]{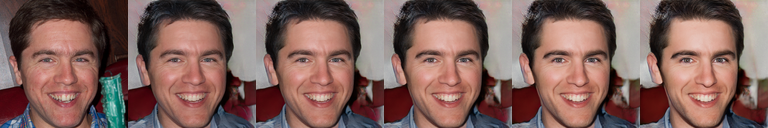}
      \includegraphics[height=2cm]{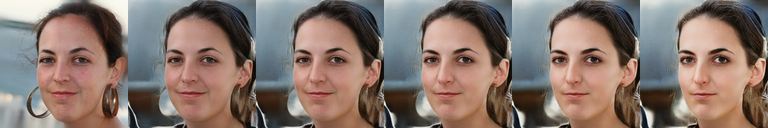}
      \includegraphics[height=2cm]{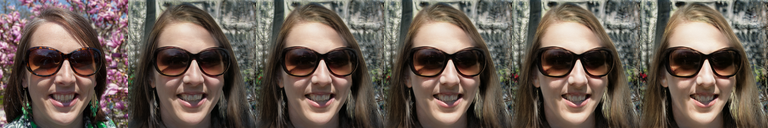}

	\end{minipage}}
	
\caption{Six-Method Results For Female, Male, Accessories Loss in Recovery Images and Eyeglasses from Top to Bottom In a Single Subfigure}
\label{fig:results}
\end{figure}

\begin{figure}[ht]
\subfloat[Beholder-GAN(Method \raisebox{.5pt}{\textcircled{\raisebox{-.9pt} {1}}})]{
	\begin{minipage}[t][4.025cm][t]{
	   1\textwidth}
	   \centering
	    \includegraphics[height=2cm]{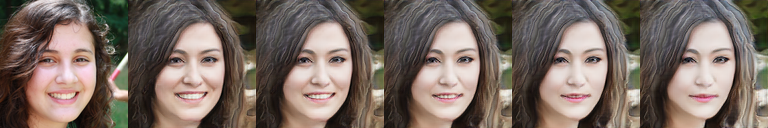}
	     \includegraphics[height=2cm]{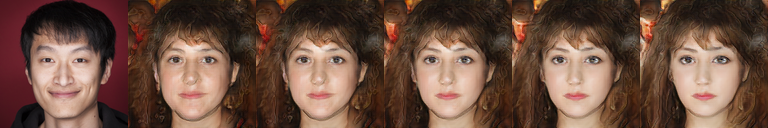}
	\end{minipage}}
 \hfill 	
   \subfloat[Method \raisebox{.5pt}{\textcircled{\raisebox{-.9pt} {3}}}]{
	\begin{minipage}[t][4.025cm][t]{
	   1\textwidth}
	   \centering
      \includegraphics[height=2cm]{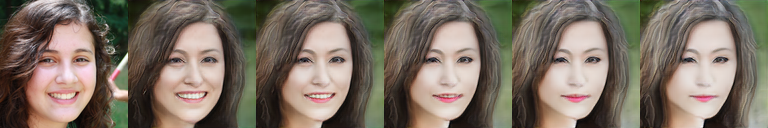}
      \includegraphics[height=2cm]{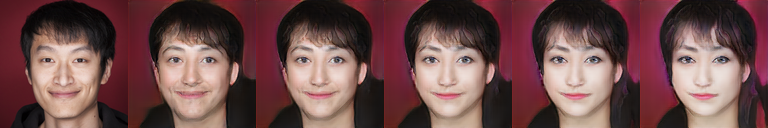}
	\end{minipage}}
	\hfill
\subfloat[Beholder-GAN(Method \raisebox{.5pt}{\textcircled{\raisebox{-.9pt} {4}}})]{
	\begin{minipage}[t][4.025cm][t]{
	   1\textwidth}
	   \centering
	    \includegraphics[height=2cm]{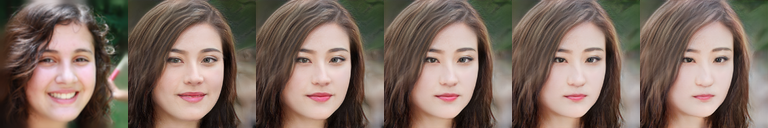}
	     \includegraphics[height=2cm]{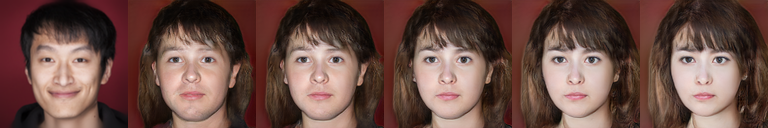}
	\end{minipage}}
 \hfill 	
   \subfloat[Method \raisebox{.5pt}{\textcircled{\raisebox{-.9pt} {6}}}]{
	\begin{minipage}[t][4.025cm][t]{
	   1\textwidth}
	   \centering
      \includegraphics[height=2cm]{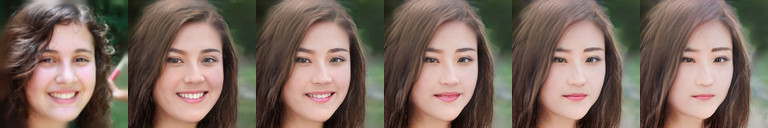}
      \includegraphics[height=2cm]{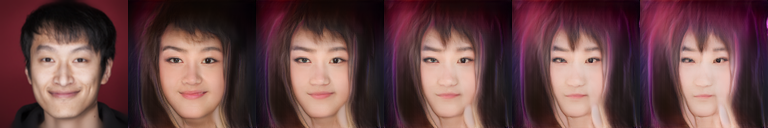}
	\end{minipage}}
	\end{figure}
	
	\newpage

\section{Conclusion and Future Work}
Our work has shown state-of-art results compared to former one, to a large extent. The discovery and research about explanation/special patterns of deep meaning of beauty embedded in GANs is intriguing, no matter which aspect we have tried to disclose, i.e. traditional latent space and functions of conditional variables. GAN-based facial attractiveness enhancement still has its possibility to make potential improvement, especially methods based on CGANs, e.g. more effective and informative identity features as conditional labels, more novel construction of CGANs for beautification, and image quality and resolution improvement of CGANs.

\clearpage

\bibliographystyle{splncs04}
\bibliography{egbib}
\end{document}